\documentclass{article}

\usepackage{microtype}
\usepackage{graphicx}
\usepackage{subfigure}
\usepackage{booktabs}
\usepackage{amsmath}
\usepackage{amssymb}
\usepackage{hyperref}

\usepackage{algorithm}
\usepackage{algorithmic}

\usepackage[accepted]{mlsys2026}
\makeatletter
\renewcommand{\Notice@String}{\textit{Preprint. Under Review.}}
\makeatother

\graphicspath{{figures/}}

\mlsystitlerunning{Working-Set Paging for Low-Resource MoE Serving}

\begin{document}

\twocolumn[
\mlsystitle{WiSP: A Working-Set View of Mixture-of-Experts Serving\\ on Extremely Low-Resource Hardware}

\begin{mlsysauthorlist}
\mlsysauthor{Jiamu Zhang}{nokia}
\mlsysauthor{Liang Wu}{nokia}
\mlsysauthor{Mayank Darbari}{nokia}
\mlsysauthor{Liangjie Hong}{nokia}
\end{mlsysauthorlist}
\mlsysaffiliation{nokia}{Nokia, Sunnyvale, CA 94085, USA}
\mlsyscorrespondingauthor{Jiamu Zhang}{jiamu.zhang@nokia.com}

\mlsyskeywords{Mixture-of-Experts, LLM serving, memory management, working set, expert offloading}

\vskip 0.3in

\begin{abstract}

Modern local and agentic workloads often need large-model capacity at low concurrency, but run on GPUs that cannot keep a frontier-scale model resident. Mixture-of-Experts (MoE) models are a natural fit because they activate only a small subset of experts per token, but their sparsity saves computation, not residency: the full expert pool still has to be stored, and any expert used by a layer must be in GPU memory when that layer runs. Static layer-level CPU offload makes such models fit, but transfers the expert layer in bulk on every forward pass, losing much of the sparsity advantage. We view low-resource MoE serving as a \emph{working-set} problem on the GPU. Routed expert weights and the KV cache are two memory-demand streams competing for the same limited VRAM. We implement this view in \textbf{WiSP} (\textbf{W}orking-\textbf{S}et \textbf{P}aging), a routing-aware expert pager that plugs into an unmodified serving engine and preserves byte-identical outputs. On a real 24\,GiB RTX 3090, WiSP achieves up to $2.0\times$ the decode throughput of static offload at the same memory budget when the model does not fit. A natural next step is to predict future experts and prefetch them. We find that this does not help in single-stream decode: the bottleneck is PCIe bandwidth, not prediction quality, so speculative transfers compete with demand transfers instead of hiding them. This shifts the design question from prefetching to allocation: how should one VRAM budget be divided between resident experts and the KV cache? We answer with \textbf{MV-WSA} (\textbf{M}arginal-\textbf{V}alue \textbf{W}orking-\textbf{S}et \textbf{A}llocation), which splits memory by marginal latency benefit per byte while enforcing a KV-admission floor. As a startup configurator, MV-WSA is the only policy we test that stays near-best on both prefill and decode; as a live controller, it resizes both pools while serving and reduces end-to-end time by up to $1.19\times$ over a fixed offline split, without changing model outputs. Our code is publicly available at \url{https://github.com/nokia-applied-research/WiSP}.

\end{abstract}
]

\printAffiliationsAndNotice{}

\section{Introduction}
\label{sec:intro}

Many language-model workloads now run outside large datacenters. Local coding assistants, on-device agents, and shared inference on fractional GPUs all need interactive latency at low concurrency, often on hardware that was not provisioned for frontier-scale models. These workloads still benefit from high-capacity models: an agent may switch between code, planning, tool use, and long-context reasoning within one session, so the deployment wants model capacity even when only one or a few requests are active.

Mixture-of-Experts (MoE) models~\citep{shazeer2017moe,fedus2022switch} are an attractive fit for this regime because they decouple capacity from per-token computation. Models such as Qwen3-30B-A3B~\citep{qwen3}, Mixtral-8$\times$7B~\citep{jiang2024mixtral}, and DeepSeek-V3~\citep{deepseekv3} place much of their parameter count in experts, but a token activates only a small routed subset. In principle, this gives a local or agentic deployment access to a large model's capacity without paying dense-model compute on every token.

The obstacle is that sparsity saves computation, but not residency. A token may use only a few experts, but the full expert pool still has to be stored somewhere, and any expert used by a layer must be in GPU memory when that layer runs. Today's MoEs have tens to hundreds of billions of parameters, so commodity GPUs cannot keep the whole model resident even though only a small fraction is active at any step.

A common way to run these models on small GPUs is offloading: keep most weights in host memory and transfer them over PCIe when the forward pass reaches them. This makes the model fit, and its simplest form is static layer-level offload, such as vLLM's \texttt{-{}-cpu-offload-gb}~\citep{kwon2023paged}. Static offload can serve a 30B MoE model in well under 10\,GiB of GPU memory, but it treats an MoE layer much like a dense one: when the layer runs, it transfers the expert weights for the layer even though routing will use only a small subset for the actual computation.

Several specialized systems improve on this baseline, but they make different tradeoffs. Some keep part of the MoE computation on the CPU, as in ktransformers, which can be fast but relies on Intel AMX~\citep{ktransformers}, while others target edge deployment by changing the weight format or precision, as in llama.cpp~\citep{llamacpp}. A third line pipelines, prefetches, or predicts expert transfers to hide communication under computation~\citep{cao2025moelightning,du2024sida,eliseev2023offload}. This last line is closest in spirit to WiSP, but its main lever is overlap: it works best when there is enough computation or batching to hide data movement. Our setting is the other end of the spectrum: in low-concurrency interactive serving, each step activates only a small routed subset of experts, but decode exposes little compute behind which to hide PCIe transfers.

The missing piece is therefore not another way to stream the next experts, but a way to decide which experts should remain resident, and how that resident set should share memory with the KV cache.

\paragraph{Serving is a working-set problem.}

In low-resource MoE serving, the scarce resources are GPU memory and PCIe bandwidth, not arithmetic. This makes the problem look much like the classical \emph{working-set} problem~\citep{denning1968ws,denning1970vm}. GPU memory is the small fast store, while host memory is the backing store, and the model generates two streams of memory demand: routed expert weights, determined by the MoE router, and KV-cache blocks, determined by the active tokens. Both have to be resident when used, and both draw from the same VRAM budget.

Under this view, good serving is memory management. The system should keep the experts a workload reuses in GPU memory, leave enough space for the KV blocks of active requests, and avoid admitting more concurrent work than those two working sets can support. This turns expert offload and KV-cache management from two separate knobs into one allocation problem: every byte given to resident experts is a byte taken away from KV, and vice versa. The same abstraction applies to any serving stack with routed experts and an autoregressive KV cache, including MoE LLMs and emerging MoE vision--language models.

The same view also shapes the system design. WiSP changes where weights live and how memory is split, not the computation itself. It leaves the serving engine, kernels, model weights, and numerical precision unchanged, so its outputs are byte-identical to the underlying engine.

WiSP implements this policy as a drop-in plug-in. It reuses the engine's existing expert-map indirection, which maps logical expert ids to physical slots, and makes those slots a small GPU scratch space containing only the resident experts. As the working set changes, WiSP updates the map rather than changing the kernel, model, or precision. This keeps the mechanism architecture-independent for any MoE model the engine can already serve. Section~\ref{sec:method} describes the mechanism in detail.

\paragraph{What we find.}

At matched GPU-memory budgets in the sub-fitting regime, WiSP's routing-aware paging improves decode throughput by up to $2.0\times$ over static offload, measured on a physically constrained 24\,GiB RTX 3090 (PCIe 4.0). This is the regime WiSP is designed for: when the model cannot be kept resident, caching only the experts a workload reuses moves far fewer bytes than static layer-level offload. Once enough VRAM is available to keep the model resident, static residency is faster because WiSP's paging indirection becomes unnecessary; we report that crossover explicitly.

We then test the natural next idea: use routing predictions to prefetch experts and hide page faults. In low-concurrency decode, this provides little benefit and can hurt, because there is too little computation behind which to hide a PCIe transfer, and thus decode is bandwidth-bound rather than prediction-bound; speculative transfers mostly compete with demand transfers for the same link. Routing information is therefore more useful as a memory signal---which experts are worth keeping resident---than as a latency signal for prefetching.

This shifts the paper's main question from prefetching to allocation. Expert weights and the KV cache share one VRAM budget: every byte given to resident experts is a byte taken away from KV, and vice versa. We introduce MV-WSA, a marginal-value allocator that splits this budget by equalizing latency benefit per byte, while enforcing a KV-admission floor so the scheduler can still admit the requested concurrency. MV-WSA can be used as a startup configurator, which fixes the split before serving, or as an online controller, which resizes the expert and KV pools while requests are running. In our real-serving allocation test, the configurator is the only policy near-best on both prefill and decode; the online controller further improves end-to-end performance by up to $1.19\times$ over the best offline split at the same VRAM budget, without changing model outputs.

We make four contributions:
\begin{itemize}
\item \textbf{A working-set formulation of low-resource MoE serving.}
We frame routed expert weights and the KV cache as two memory-demand streams competing for the same limited VRAM. This gives a common vocabulary for expert residency, KV allocation, thrashing, and admission, and separates our low-concurrency setting from high-batch datacenter serving (Section~\ref{sec:bg}).

\item \textbf{A drop-in routing-aware expert pager.}
We implement expert paging through the serving engine's existing expert-map indirection, without changing kernels, model weights, or quantization. On a physically constrained 24\,GiB RTX 3090, WiSP improves decode throughput by up to $2.0\times$ over static offload at matched VRAM when the model does not fit (Sections~\ref{sec:method},~\ref{sec:finding}).

\item \textbf{A negative result on prefetching in low-concurrency decode.}
We show that routing-based expert prefetching is not the main speed lever in this regime: single-stream decode is PCIe-bandwidth-bound, not prediction-bound, so speculative transfers compete with demand transfers instead of hiding them. The routing signal is more useful for sizing the resident expert set than for reducing latency through prefetching (Section~\ref{sec:finding}).

\item \textbf{MV-WSA for joint expert--KV allocation.}
We introduce a marginal-value allocator that divides VRAM between resident experts and the KV cache under an admission floor. As a startup configurator, it is near-best on both prefill and decode in our real-serving allocation test. As an online controller, it physically resizes both pools during serving and improves end-to-end performance by up to $1.19\times$ over the best offline split at the same VRAM budget, byte-identically (Sections~\ref{sec:method},~\ref{sec:eval}).
\end{itemize}

\section{Background: Serving as a Working-Set Problem}
\label{sec:bg}

\subsection{The Working-Set View}

A program's \emph{working set} is the set of memory pages it has used recently~\citep{denning1968ws,denning1970vm}. If that set stays resident, execution is fast, but if the resident set is too small, the program repeatedly faults on pages it just used and begins to \emph{thrash}. If the system admits more concurrent work than the combined working sets can support, the whole machine thrashes.
Low-resource MoE serving has the same shape: each active token needs a routed subset of experts in every MoE layer, and each active request needs its KV blocks. Both must be resident in GPU memory when used. As concurrency, context length, or the routed expert union grows, the two working sets compete for the same limited VRAM.

This view is useful because expert use is sparse, but it is not structureless. A token activates only $k$ of $N$ experts, so keeping every expert resident wastes memory when VRAM is limited. At the same time, a short sequence of decode steps touches much more than a tiny per-token set, so the cache cannot shrink to just the next token's experts. The opportunity is reuse at the workload scale: across bursts and sessions, a workload tends to revisit a stable subset of experts. Caching that routed working set, rather than the whole expert pool, is what makes expert paging viable on a device that cannot keep the full model resident.

The KV cache is the other half of this memory problem. PagedAttention~\citep{kwon2023paged} already manages KV as blocks that can be allocated, reused, and spilled. Expert weights, however, are usually handled by a separate mechanism: either kept fully resident or streamed in bulk from the host. The two mechanisms still draw from the same VRAM budget. Every byte used for resident experts is a byte unavailable to KV, and vice versa. Seen this way, low-resource MoE serving is not two independent problems---expert offload and KV management---but one allocation problem over two competing working sets.

\subsection{Why the Serving Regime Matters}

The best strategy depends on concurrency. In high-batch datacenter serving, a layer may activate many or most experts across the batch. The per-layer working set is then close to the whole expert bank, and there is enough computation to overlap expert transfers with execution. Streaming and prefetching are natural in that setting~\citep{liu2026fluxmoe,cao2025moelightning}.

WiSP targets the other end of this spectrum: one or a few interactive streams on a GPU that cannot hold the full model. Here each step activates only a small routed subset of experts, and decode exposes little computation behind which to hide PCIe transfers. The useful question is therefore not how to hide the next transfer, but which experts should remain resident across steps and how much VRAM should be left for KV. Section~\ref{sec:finding} measures this directly.

\subsection{Scope and Generality}
\label{sec:scope}

The expert-paging mechanism applies to any model with routed experts that the serving stack presents as an expert bank. The full working-set allocation problem applies when the model also has an autoregressive KV cache, because resident experts and KV blocks compete for the same memory budget. Models without a KV cache, such as diffusion MoEs or pure recurrent SSM--MoE variants, can still use expert paging to reduce resident memory. They do not exercise the full expert--KV allocation problem, so we treat them as mechanism-generalization cases rather than headline throughput targets. Table~\ref{tab:scope} summarizes this distinction.

\begin{table*}[t]
\caption{Where the working-set view applies. \emph{Expert stream} = routed experts that can be paged;
\emph{KV stream} = an autoregressive key--value cache that co-contends for memory. WiSP's mechanism
needs the former; its full theory needs both.}
\label{tab:scope}
\vskip 0.1in
\begin{center}
\scalebox{0.7}{
\begin{tabular}{lllccl}
\toprule
Architecture & MoE variant & Representative & Expert stream & KV stream & WiSP coverage \\
\midrule
Dense LLM & $\times$ (dense) & Llama-3, Qwen2.5 & $\times$ & \checkmark & n/a \\
SSM (pure) & \checkmark & BlackMamba~\citep{anthony2024blackmamba} & \checkmark & $\times$ (recurrent state) & mechanism only \\
\textbf{MoE LLM} & \checkmark & Mixtral, Qwen3-MoE, DeepSeek-V3 & \checkmark & \checkmark & \textbf{full (primary)} \\
\textbf{MoE VLM} & \checkmark & DeepSeek-VL2, Kimi-VL, Aria & \checkmark & \checkmark & \textbf{full (breadth)} \\
Hybrid SSM--MoE & \checkmark & Jamba~\citep{lieber2024jamba} & \checkmark & \checkmark (attention) & \textbf{full} \\
MoE diffusion LLM & \checkmark & LLaDA-MoE~\citep{zhu2025lladamoe} & \checkmark & $\times$ (bidirectional) & mechanism only \\
MoE image diffusion & \checkmark & DiT-MoE~\citep{fei2024ditmoe} & \checkmark & $\times$ (latent denoise) & mechanism only \\
\bottomrule
\end{tabular}
}
\end{center}
\vskip -0.1in
\end{table*}

\paragraph{Where WiSP sits among existing systems.}
Most prior MoE-offloading work targets autoregressive MoE LLMs. The main axis separating these systems is the serving regime. High-batch systems can stream or prefetch experts because they have enough computation to overlap transfers. WiSP targets low-concurrency serving, where overlap is limited and reuse-based caching matters more. It changes residency and allocation policy, not the model computation, so the engine's outputs remain byte-identical.

The closest concurrent system is FluxMoE~\citep{liu2026fluxmoe}. Like WiSP, it trades expert residency against KV capacity, but it uses a stream-and-evict policy tuned for high-batch overlap. WiSP instead uses cache-and-reuse for the regime where there is little compute to hide transfer latency. Beyond autoregressive LLMs, MoE VLMs, hybrid models, and diffusion MoEs use different serving stacks; WiSP attaches to them either as the same serving stack plug-in or through a small wrapper. Appendix~\ref{app:designspace} gives the full design-space and serving-stack comparisons (Tables~\ref{tab:design},~\ref{tab:stacks}).

\section{WiSP}
\label{sec:method}

\begin{figure*}[t]
\centering
\includegraphics[width=\textwidth]{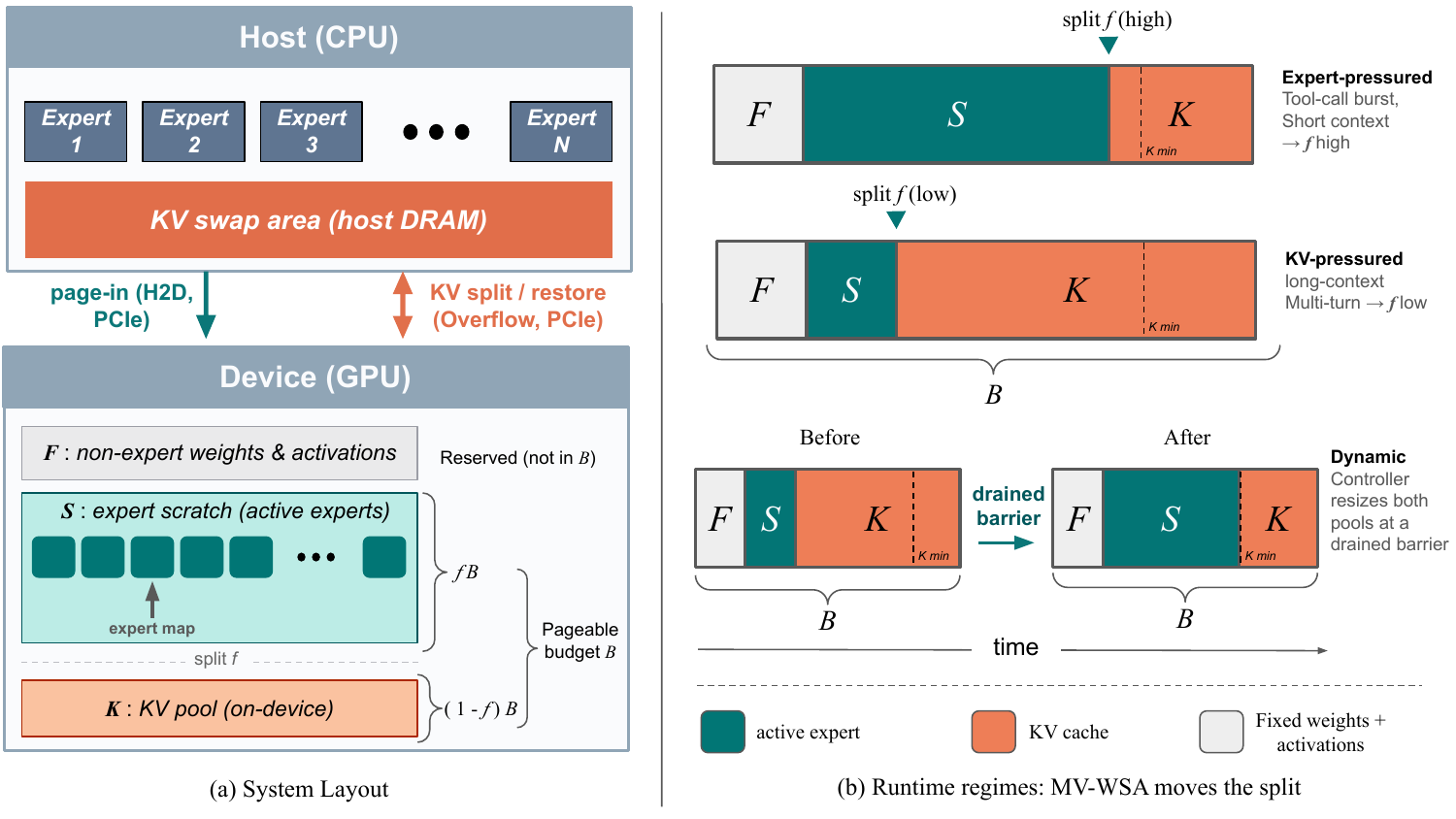}
\vspace{-4mm}
\caption{WiSP at a glance. \textbf{(a) System layout.} The full expert weights stay pinned on the host;
WiSP \emph{pages in} only the routed experts to a small GPU scratch $S$ over PCIe and runs the
unmodified fused-MoE kernel through the \texttt{expert\_map} $\pi$, so output is byte-identical. The
pageable budget $B = B_{\mathrm{exp}} + B_{\mathrm{kv}}$---the expert scratch $S$ plus the KV pool $K$,
\emph{excluding} the fixed non-expert slab $F$---is divided by the split $f$; KV that overflows the
on-device pool spills to host. \textbf{(b) Runtime regimes.} The optimal split moves with the
workload---expert-heavy for tool-call bursts, KV-heavy for long contexts---and MV-WSA tracks it, in the
dynamic case by physically resizing both pools at a drained barrier, byte-identically.}
\label{fig:overview}
\end{figure*}

WiSP separates low-resource MoE serving into two decisions. First, a \emph{routing-aware expert pager} (\S\ref{sec:method-pager}) decides which experts are resident in GPU memory. It uses the serving engine's existing \texttt{expert\_map} indirection to map routed expert ids onto a small GPU scratch space, so the kernel, weights, precision, and outputs are unchanged.
Second, \emph{MV-WSA} (Marginal-Value Working-Set Allocation, \S\ref{sec:method-split}--\S\ref{sec:method-dynamic}) decides how much of the available VRAM should go to that expert scratch versus the KV cache. The pager manages residency \emph{within} the expert pool; MV-WSA moves the boundary \emph{between} the expert and KV pools. We use MV-WSA in two modes: offline, as a startup configurator that fixes the split before serving, and online, as a controller that resizes both pools while requests are running. Figure~\ref{fig:overview} summarizes the design.

\subsection{Setting and Notation}
\label{sec:method-setting}

The model has $L$ MoE layers with $N$ experts each, and the router selects $k \ll N$ experts per token per layer. At step $t$, layer $\ell$ selects
$R_\ell^{(t)} \subseteq \{1,\dots,N\}$, and every expert $e \in R_\ell^{(t)}$ must be resident when that layer executes.
After fixed GPU costs are reserved---non-expert weights, activations, CUDA context, and other runtime state---the device has a pageable budget $B$ shared by resident experts and the KV cache:
\begin{equation}
\label{eq:split}
B \;=\; B_{\mathrm{exp}} + B_{\mathrm{kv}}, \qquad
f \;\triangleq\; B_{\mathrm{exp}}/B \;\in\; (0,1).
\end{equation}
The split $f$ is MV-WSA's control variable. If one expert occupies $M_e$ bytes, then increasing the per-layer resident cap by one expert costs $a \triangleq L M_e$ bytes across all MoE layers, so the expert budget supports
$C = B_{\mathrm{exp}}/a$ resident experts per layer. The KV budget supports
$\kappa = B_{\mathrm{kv}}/b$ KV blocks, where $b$ is the GPU memory per KV block summed over all layers. At a fixed budget $B$, WiSP's allocation problem is to choose where this split should land.

\subsection{The Expert Pager}
\label{sec:method-pager}

The allocator can only trade expert memory against KV memory if the expert side is itself pageable. WiSP provides this substrate with the engine's existing expert-map indirection, without changing the fused-MoE kernel, model weights, or precision.

\paragraph{Expert-map indirection.}
Modern fused-MoE kernels accept an \texttt{expert\_map} $\pi$ that maps logical expert ids to physical slots. WiSP makes those physical slots a small GPU scratch space with capacity $C$ per layer. The full expert weights remain in host memory; at each layer, WiSP ensures that the routed experts are present in scratch, updates $\pi$, and dispatches the unmodified kernel. A page fault occurs when a routed expert is not currently resident, in which case WiSP copies it from the host master and evicts an older resident expert using LRU. All dynamic behavior is in the map and the scratch contents, not in the computation.
\paragraph{Per-step protocol.}
Algorithm~\ref{alg:wisp} shows the per-layer decode step. Given the routed expert set $R$, WiSP copies only the missing experts, assigns them scratch slots, refreshes the LRU state for the experts used by this step, and calls the stock fused-MoE kernel with the updated expert map.
\begin{algorithm}[t]
   \caption{\textsc{WispLayer.Step}($R$) at one MoE layer}
   \label{alg:wisp}
\small
\begin{algorithmic}[1]
   \STATE {\bfseries Input:} needed experts $R$; cap $C$; maps $\pi,\pi^{-1}$; LRU ticks $\tau$; clock $c$; CPU master $W^{\mathrm{cpu}}$; GPU scratch $S$; device map $\pi_{\mathrm{dev}}$
   \STATE $M \gets \{e \in R : \pi(e) = \bot\}$ \hfill$\triangleright$ faults
   \STATE $U \gets \{s : \pi^{-1}(s) \notin R\}$ in increasing $\tau$ \hfill$\triangleright$ LRU victims
   \STATE \textsc{assign} each $e \in M$ a victim slot $s_e$ from $U$
   \FOR{each $e \in M$ with slot $s_e$}
     \STATE $S[s_e] \gets \textsc{copy\_async}(W^{\mathrm{cpu}}[e])$
     \STATE $\pi(e)\gets s_e;\ \pi^{-1}(s_e)\gets e;\ \pi_{\mathrm{dev}}[e]\gets s_e$
   \ENDFOR
   \STATE $c \gets c+1$;\quad \textbf{for} $e \in R$ \textbf{do} $\tau[\pi(e)] \gets c$
   \STATE \textbf{return} $\textsc{FusedMoE}(\mathrm{hidden},\, W{=}S,\, \texttt{expert\_map}{=}\pi_{\mathrm{dev}})$
\end{algorithmic}
\end{algorithm}
The key invariant is exactness. When the kernel runs, every routed expert id maps to a scratch slot containing the same bytes the full expert tensor would have supplied. The kernel is unchanged, so WiSP's output matches the unpaged engine up to the engine's own floating-point reproducibility. For large prefill batches whose expert union exceeds $C$, WiSP groups tokens so that each group fits the scratch and reassembles the outputs in token order; this path is an implementation detail and does not trigger in our single-stream decode measurements.

\paragraph{Reuse-distance miss curves.}
WiSP uses LRU because the low-concurrency decode regime is reuse-limited rather than overlap-limited: each step has too little computation to hide a PCIe transfer, so the useful policy is to keep the experts a workload comes back to. LRU also gives a simple way to estimate the value of more memory. From a reference trace, the reuse-distance histogram determines the miss curve $m(C)$: how many references would miss at resident cap $C$. We use this miss-curve view for the expert stream, and use the analogous block-level view for KV, with the admission floor below capturing the scheduler cliff that a smooth miss curve misses. 

\subsection{The Expert$\leftrightarrow$KV Split}
\label{sec:method-split}

A split $f$ gives $C = fB/a$ expert slots per layer and $\kappa = (1-f)B/b$ KV blocks. The allocator asks a simple question: if we move one byte from KV to experts, do we remove more latency than if we leave that byte in KV?

Let $m_{\mathrm{exp}}(C)$ be the expert miss count at cap $C$, and let $m_{\mathrm{kv}}(\kappa)$ be the KV miss or recompute cost induced by a KV pool of $\kappa$ blocks. Expert misses and KV misses have different costs. An expert miss transfers weights over PCIe, with cost $c_{\mathrm{exp}} = M_e t_{\mathrm{byte}}$. A KV miss costs recomputation, with cost $c_{\mathrm{kv}} = \beta b_{\mathrm{tok}}$. The latency model at split $f$ is
\begin{equation}
\label{eq:latency}
T(f) \;=\; c_{\mathrm{exp}}\, m_{\mathrm{exp}}(C) \;+\; c_{\mathrm{kv}}\, m_{\mathrm{kv}}(\kappa).
\end{equation}

The best split depends on the workload. Short tool-call bursts have little KV pressure and benefit from keeping more reused experts resident, so the optimum shifts expert-heavy. Long-context or multi-turn workloads put more pressure on KV, so the optimum shifts back toward KV. At an interior optimum, the split is equimarginal: one more byte should buy the same latency reduction in either pool,
\begin{equation}
\label{eq:equimarginal}
c_{\mathrm{exp}}\, m'_{\mathrm{exp}}(C)/a \;=\; c_{\mathrm{kv}}\, m'_{\mathrm{kv}}(\kappa)/b .
\end{equation}
MV-WSA follows this principle: move bytes toward whichever pool currently removes more latency per byte.

\paragraph{The admission floor.}
The smooth cost in Eq.~\eqref{eq:latency} misses one hard constraint. A continuous-batching scheduler must have enough KV capacity to admit the active requests. If the GPU KV pool is too small, the scheduler may preempt and re-prefill repeatedly, or the engine may refuse to initialize. This is a cliff, not a gradual miss penalty. We therefore enforce
\begin{equation}
\label{eq:admission}
\kappa \;\ge\; \kappa_{\min} \;=\; W \cdot \lceil \ell_{\mathrm{ctx}}/b_{\mathrm{tok}} \rceil ,
\end{equation}
where $W$ is the served concurrency and $\ell_{\mathrm{ctx}}$ is the per-session context length. The floor is slack when KV is the pressured pool. It becomes binding when the unconstrained optimum pushes the split toward experts: $\kappa_{\min}$ is what keeps an expert-heavy split admissible. This is the difference between MV-WSA and a KV-blind expert-proportional split, which we measure as a startup failure in Section~\ref{sec:eval-mvwsa}.

\subsection{MV-WSA: Configurator and Online Controller}
\label{sec:method-mvwsa}

MV-WSA does not need the full miss curves in advance. It builds them from decayed reuse-distance histograms while the workload runs, then moves the split toward the equimarginal point in Eq.~\eqref{eq:equimarginal}. A small hysteresis band prevents the split from chasing noise. We use the same allocation principle in two modes.

In \emph{offline} mode, MV-WSA runs on a short representative trace and returns a fixed split $\hat f$ used at startup. Given a budget $B$, it emits the two engine knobs that realize this split: the per-layer expert cap and the GPU KV pool, subject to the admission floor,
\begin{equation}
\label{eq:config}
C = \min\!\bigl( \lfloor \hat f B/a \rfloor,\ \lfloor (B - \kappa_{\min} b)/a \rfloor \bigr),
\qquad B_{\mathrm{kv}} = B - C a .
\end{equation}
All other memory settings are held fixed, so comparisons are iso-VRAM: the arms differ only in where the same bytes go.

In \emph{online} mode, MV-WSA updates the split while serving. A single startup split is only the best fixed compromise over a trace, but an agentic session can move between regimes: long-context turns may need KV, while short tool-call bursts may leave KV underused and benefit from more resident experts.

\subsection{Dynamic Dual-Resize}
\label{sec:method-dynamic}

The live controller implements the online case using a simple observation from the engine: once the KV pool covers the current working set, stays above $\kappa_{\min}$, and has a small amount of headroom, extra KV blocks have near-zero marginal value. The controller therefore measures the peak KV occupancy at a drained barrier, keeps KV just above that peak, and gives the remaining bytes to the expert scratch. In this regime, the equimarginal rule becomes simple: once KV has enough room, the next byte is worth more on the expert side. Appendix~\ref{app:dynamic} gives the full control law.

Applying the new split inside a running engine requires two resize operations: resizing the per-layer expert scratch and resizing the vLLM KV-block pool. Both happen only at a drained barrier, when no request owns a block being moved. The controller frees the shrinking side before growing the other, so the transient footprint never exceeds the fixed budget $B$. The result is still an iso-VRAM comparison, in which the controller only moves bytes from one pool to the other. Because it moves stored tensors rather than changing computation, greedy outputs are preserved. If a resize fails, the engine keeps the previous valid split.

\subsection{Implementation}
\label{sec:method-impl}

WiSP is implemented as a drop-in plug-in over an unmodified vLLM. Expert masters live in host memory, and each MoE layer owns a GPU scratch whose size is set by the expert cap $C$. The static configurator realizes a split by setting the expert cap and the on-device KV pool; the online controller adds resize operations for both pools and applies them only at drained barriers. Appendix~\ref{app:impl} gives the engineering details.

\section{The Central Finding: Prediction Is a Memory Lever, Not a Speed Lever}
\label{sec:finding}

The working-set view suggests a natural optimization: if we can predict which experts a token will need, we should prefetch them and hide page faults. This is not a strawman. Routing is partly predictable because later-layer routing depends on hidden states produced by earlier experts, and our co-activation predictor reaches nontrivial next-expert precision (Appendix~\ref{app:predictor}). The question is whether that prediction turns into decode speed.

\begin{figure*}[t]
\centering
\includegraphics[width=0.92\textwidth]{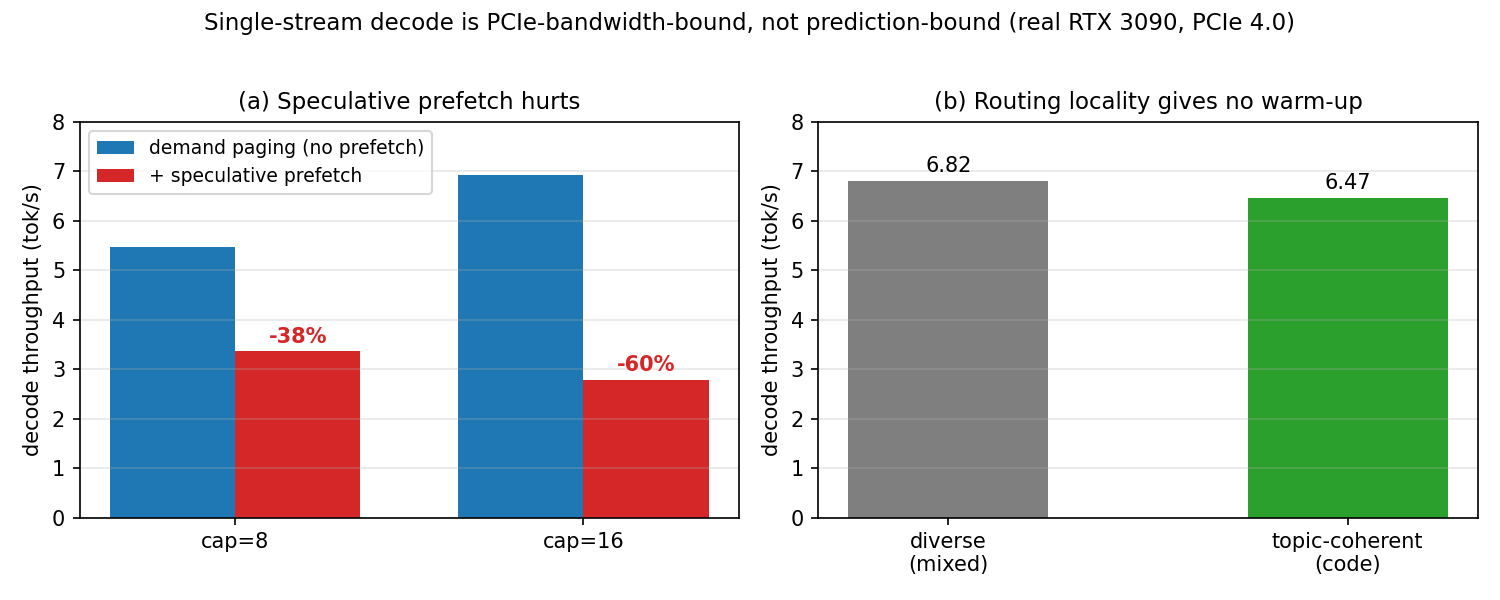}
\caption{Prediction does not buy decode speed at single-stream, measured on a real 24\,GiB RTX 3090 (PCIe 4.0). (a) Turning on speculative prefetch with the online co-activation predictor \emph{lowers} decode throughput by 38--60\% at constrained caps (cap 8: $5.47\to3.37$; cap 16: $6.94\to2.79$ tok/s) and raises time-to-first-token by $1.6$--$1.9\times$ (Qwen3-30B-A3B). (b) A topic-coherent (code) session reaches no higher steady-state throughput than a deliberately diverse one ($6.47$ vs.\ $6.82$ tok/s), so there is no routing-locality warm-up to harvest. The binding constraint is PCIe bandwidth, not prediction quality.}
\label{fig:negative}
\end{figure*}

In the low-concurrency regime, it does not. Figure~\ref{fig:negative} tests speculative expert prefetching on Qwen3-30B-A3B on a real 24\,GiB RTX 3090 (PCIe 4.0). Turning on prefetch lowers decode throughput by 38--60\% at constrained caps and raises time-to-first-token by $1.6$--$1.9\times$. A second test compares a topic-coherent session with a deliberately diverse one; despite having more routing locality, the coherent session reaches no higher steady-state throughput ($6.47$ vs.\ $6.82$ tok/s). There is no warm-up benefit to harvest.

The reason is bandwidth. At batch-1 decode, each layer computes a single token, so the MoE GEMM is too small to hide a side-stream PCIe transfer. Speculative copies therefore compete with demand copies for the same link, and at tight caps they can evict experts that the next step still needs. The measured traffic makes the limit concrete: on Qwen3-30B-A3B, only 44--48\% of each layer's top-8 experts carry over between adjacent decode steps. Even a perfect last-step cache must still stream about 1.8\,GiB of expert weights per token. On our RTX 3090, with about 25\,GB/s host-to-device bandwidth, that mandatory traffic alone caps decode at roughly 13--16 tok/s. The pipe is already full; prediction can reshuffle traffic, but it cannot widen the pipe.

This result changes how we use routing information. In this regime, routing is not a latency signal for prefetching; it is a memory signal for deciding how large the resident expert set should be. Once expert paging frees part of the model from GPU residency, the next question is how to spend that freed memory: keep more experts resident, or give the bytes to the KV cache? That allocation problem is what MV-WSA solves (\S\ref{sec:method-split}--\S\ref{sec:method-dynamic}) and what we evaluate in \S\ref{sec:eval-mvwsa}.

\section{Evaluation}
\label{sec:eval}

The evaluation follows the paper's argument. First, WiSP should not change the model's computation, so we check output preservation (Section~\ref{sec:eval-correct}). Second, if expert residency is a working-set problem, routing-aware paging should move less data than static layer-level offload at the same GPU budget; we test this on a physically constrained 24\,GiB RTX 3090 (Section~\ref{sec:eval-isovram}). Third, because expert memory and KV memory share one store, we evaluate whether MV-WSA chooses a better split than fixed or memory-blind policies, both as a startup configurator and as a live resize controller (Section~\ref{sec:eval-mvwsa}).

\paragraph{Experimental setup and scope.}
The headline iso-VRAM throughput results for Qwen3-30B-A3B and Kimi-VL-A3B (Figure~\ref{fig:isovram}, Table~\ref{tab:main}) are measured on a physically constrained consumer GPU: a single 24\,GiB NVIDIA RTX 3090 on a real PCIe~4.0 x16 link (measured host-to-device bandwidth ${\approx}25$\,GB/s), vLLM 0.11.2, eager mode, temperature 0. Both models exceed 24\,GiB in BF16 (Qwen3 ${\approx}57$\,GiB, Kimi-VL ${\approx}31$\,GiB), so paging is forced by the hardware rather than emulated. The H100 results serve different purposes: MiniMax-M2 and Jamba are capability checks on models that exceed one 94\,GiB H100, while the OLMoE MV-WSA startup-configurator test isolates the allocation policy under a controlled 8\,GiB budget. The diffusion experiments use a standalone PyTorch pager, so we report correctness and resident-memory behavior there, not serving latency.

\subsection{Paging Is Output-Preserving}
\label{sec:eval-correct}

At temperature 0, WiSP matches the unmodified engine for every cap we test, and this is the expected consequence of property P1: the fused-MoE kernel is unchanged, and every routed expert is read from byte-equal weights. We also check the mechanism on diffusion models, where the comparison is stricter because we can inspect the whole generation trajectory: DiT-MoE latents match with maximum absolute difference zero across the cap sweep, and LLaDA-MoE completions are token-identical (Section~\ref{sec:eval-general}).

\subsection{Iso-VRAM Decode Throughput versus Static Offload}
\label{sec:eval-isovram}

Table~\ref{tab:main} summarizes the main system cases. Qwen3 and Kimi-VL are matched-budget throughput comparisons on the 24\,GiB edge GPU. Jamba and MiniMax-M2 are capability checks on models whose native serving paths fail or become impractical under the stated memory constraints. The diffusion mechanism-only cases have no serving-throughput baseline and are reported in Appendix~\ref{app:extended-eval}.

\begin{table*}[t]
    \caption{Main results across the full-system cases. Where a throughput comparison is possible we match GPU-memory budgets: \emph{naive offload} is vLLM's static \texttt{-{}-cpu-offload-gb} (offloaded weight bytes, in GiB) and \emph{WiSP paging} keeps a cap-sized resident expert scratch (cap, in experts/layer); both columns report decode \emph{tok/s} at that budget. \textsc{edge} = measured on a physically constrained 24\,GiB RTX 3090. \textsc{real}/\textsc{cap.} = the model exceeds one 94\,GiB H100 or hits an unsupported native path, so the native baseline does not run ($\times$); these rows are capability checks rather than throughput claims and have no speedup (``---'').}
    \label{tab:main}
    \vskip 0.1in
    \begin{center}
    \footnotesize
    \setlength{\tabcolsep}{5pt}
    \begin{tabular}{@{}llrrrrc@{}}
    \toprule
     & & \multicolumn{2}{c}{naive offload} & \multicolumn{2}{c}{WiSP paging} & \\
    \cmidrule(lr){3-4}\cmidrule(lr){5-6}
    Setting & GPU budget & GiB & tok/s & cap & tok/s & Speedup \\
    \midrule
    \multicolumn{7}{@{}l}{\emph{MoE LLM --- Qwen3-30B-A3B (BF16), 128 experts, top-8, on a 24\,GiB RTX 3090 / PCIe 4.0 (cf.~Figure~\ref{fig:isovram})}} \\
    \textsc{edge} & ${\approx}20$\,GiB   & 50 & 4.17 & 16 & 7.62          & $1.83\times$ \\
    \textsc{edge} & ${\approx}22$\,GiB   & 44 & 4.93 & 32 & \textbf{9.93} & $\mathbf{2.01\times}$ \\
    \midrule
    \multicolumn{7}{@{}l}{\emph{MoE VLM --- Kimi-VL-A3B (BF16), 64 experts, top-8, on a 24\,GiB RTX 3090 / PCIe 4.0}} \\
    \textsc{edge} & ${\approx}14.7$\,GiB & 20 & 5.00 & 16 & 8.97 & $1.79\times$ \\
    \textsc{edge} & ${\approx}8.8$\,GiB  & 24 & 4.23 & 8  & 6.69 & $1.58\times$ \\
    \midrule
    \multicolumn{7}{@{}l}{\emph{Hybrid SSM--MoE --- Jamba-v0.1 (52B, 103\,GiB BF16), 16 experts, top-2}} \\
    \textsc{real} & 94\,GiB H100 & \multicolumn{2}{c}{full residency: $\times$ OOM ($103{>}94$)}          & 4 & 5.73 & --- \\
    \textsc{real} & 94\,GiB H100 & \multicolumn{2}{c}{\texttt{cpu-offload-gb} (any): $\times$ assertion} & 8 & 9.80 & --- \\
    \midrule
    \multicolumn{7}{@{}l}{\emph{MoE LLM --- MiniMax-M2 (229B FP8), 256 experts, top-8}} \\
    \textsc{cap.} & 94\,GiB H100 & \multicolumn{2}{c}{offload ${\geq}180$: $\times$ host OOM (256\,GiB cgroup)} & 32 & 3.70\,{\scriptsize(14\,GiB exp.)} & --- \\
    \bottomrule
    \end{tabular}
    \end{center}
    \vskip -0.1in
    \end{table*}
    
\begin{figure*}[t]
\centering
\includegraphics[width=0.6\textwidth]{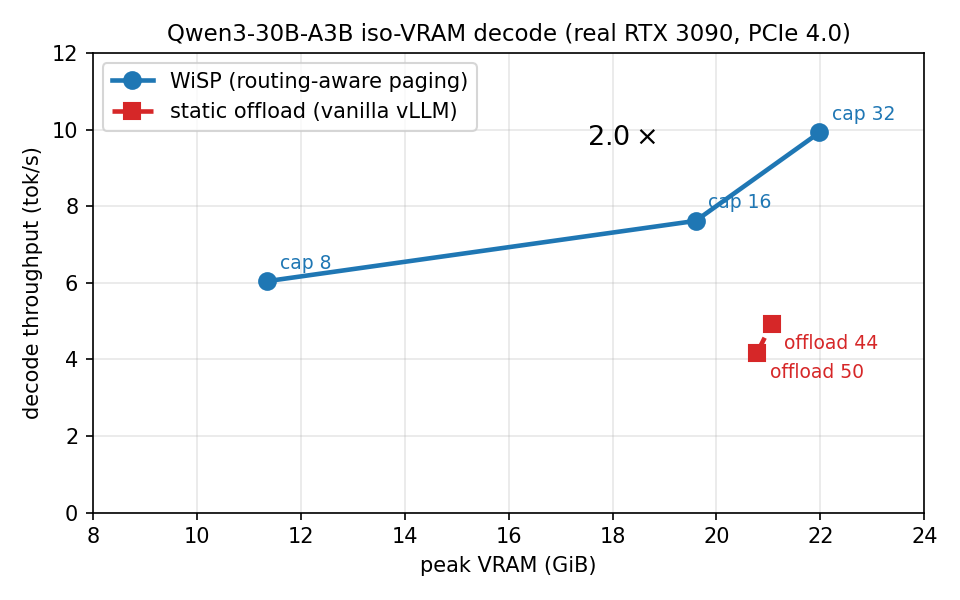}
\caption{Iso-VRAM decode throughput on Qwen3-30B-A3B, measured on a real 24\,GiB RTX 3090 (PCIe 4.0).
When the model does not fit, routing-aware paging (WiSP) moves fewer bytes per step than static
layer-level offload. At a matched ${\approx}22$\,GiB budget, WiSP decodes ${\approx}2.0\times$ faster
(cap 32: 9.93 vs.\ offload 44: 4.93 tok/s). Within the range this card can hold, increasing the expert
cap improves WiSP throughput (cap 8/16/32 $\to$ 6.0/7.6/9.9 tok/s), while the routing-blind baseline
remains bandwidth-limited. A 24\,GiB card cannot reach the full-footprint budget where the model fits
and static residency wins; WiSP targets the sub-fitting regime, not a universal speedup.}
\label{fig:isovram}
\end{figure*}

Figure~\ref{fig:isovram} is the main systems result on real edge hardware. On the 24\,GiB RTX 3090, both WiSP and the static-offload baseline run under the same GPU budget, but they spend the bytes differently. Static offload streams the expert layer in bulk, so its decode throughput stays near 4--5 tok/s. WiSP keeps a routed expert working set resident, so throughput rises as the cap grows: 6.0, 7.6, and 9.9 tok/s for cap 8, 16, and 32. The result is a ${\approx}1.8$--$2.0\times$ speedup in the sub-fitting range this card can actually exercise. A 24\,GiB card cannot reach the full-residency crossover; we only claim the regime where the model does not fit.

Kimi-VL gives a second check on the same mechanism in a different model family. It is a MoE vision--language model with a dense vision tower and a routed MoE language backbone, yet WiSP uses the same vLLM plug-in without model-specific code. At matched budgets, it improves decode throughput by $1.58$--$1.79\times$ over static offload (Table~\ref{tab:main}) while preserving outputs on both text and image prompts (Section~\ref{sec:eval-general}).

\paragraph{Scaling to MiniMax-M2: a capability check.}
We also run WiSP on the 229B-parameter FP8 MiniMax-M2 on one 94\,GiB H100. This is a capability check, not a throughput claim. WiSP serves the model with coherent output while keeping only a $7$--$14$\,GiB expert scratch on the GPU. Stock vLLM can also serve the model on this card with enough static offload, and is faster, but lowering its GPU footprint requires more offloaded and pinned/staged host memory until host RAM becomes the bottleneck. WiSP's point here is different: the full expert set lives once in host memory, while the GPU holds only the resident scratch. A clean iso-GPU-VRAM comparison where this tradeoff pays off requires a physically smaller card, which we leave to future hardware work.

The throughput result is only the expert side of the trade. The working-set view says that the expert cache and the KV cache share one store: shrinking the resident expert cap leaves more GPU memory for on-device KV, while growing the expert cap reduces the KV pool available to the scheduler. Push the expert cap too high and admission can fail because the engine has no usable KV pool. Table~\ref{tab:mvwsa-dynamic} measures the reverse move online, reclaiming idle KV capacity and giving those bytes back to resident experts. This is the allocation problem MV-WSA is designed to solve.

\subsection{Allocating the Budget: MV-WSA}
\label{sec:eval-mvwsa}

MV-WSA chooses the split between resident experts and on-device KV. We evaluate it in two settings. First, as a static startup configurator, it chooses a fixed split before serving and we compare that split against fixed and KV-blind policies on real \texttt{vllm serve} (Table~\ref{tab:mvwsa-live}). Second, as a live controller, it physically moves bytes between the expert scratch and the KV pool while requests are running (Table~\ref{tab:mvwsa-dynamic}).

\paragraph{Static configurator, real serving.}
We run one \texttt{vllm serve} instance per arm and replay \texttt{os} sessions from AgentInstruct~\citep{zeng2024agenttuning}, the agent-trajectory dataset whose \texttt{os}/\texttt{db}/\texttt{webshop} splits are collected on AgentBench environments~\citep{liu2024agentbench}, at concurrency~4. Each session is prepended with a distinct ${\sim}1.2$--$1.5$k-token pseudo-document, so concurrent requests hold separate KV state rather than sharing a cached prefix. All arms use the same total GPU budget and host CPU-swap; only the expert--KV split changes.

We compare four policies: \textbf{MV-WSA}, which uses the configurator split $\hat f$ subject to the admission floor; \textbf{50--50}, an even split of the pageable pool; \textbf{KV-blind prop.}\footnote{This is a deliberately KV-blind prior, not FluxMoE~\citep{liu2026fluxmoe}, which targets high-batch stream-and-evict overlap.}, an expert-working-set-proportional split that ignores KV demand; and \textbf{naive offload}, vanilla \texttt{-{}-cpu-offload-gb} with WiSP off. Table~\ref{tab:mvwsa-live} runs this on OLMoE-1B-7B at an 8\,GiB budget, a KV-pressured regime where wasting KV capacity shows up immediately in TTFT.

\begin{table}[t]
\caption{MV-WSA startup configurator (\emph{static} split, fixed for the run) on real \texttt{vllm serve}, iso-VRAM, AgentInstruct~\citep{zeng2024agenttuning} \texttt{os}, concurrency~4, OLMoE-1B-7B at an 8\,GiB budget. Every arm gets the same total GPU budget and host CPU-swap; only the expert--KV split changes. \emph{Turn wall} is median wall-clock time per turn (prefill$+$decode), lower is better.}
\label{tab:mvwsa-live}
\vskip 0.1in
\begin{center}
\scalebox{0.78}{
\begin{tabular}{@{}llcccc@{}}
\toprule
Arm (split) & cap & KV (GiB) & TTFT med & decode & turn wall (s) \\
\midrule
\textbf{MV-WSA} & 21 & 2.6 & \textbf{2.9} & 40.6 & \textbf{6.3} \\
50--50          & 18 & 3.2 & 2.9 & 29.2 & 7.0 \\
KV-blind prop.\ & 35 & 0.5 & 19.8 & \textbf{56.2} & 22.0 \\
naive offload   & --- & --- & 21.3 & 30.7 & 25.5 \\
\bottomrule
\end{tabular}
}
\end{center}
\vskip -0.1in
\end{table}

Table~\ref{tab:mvwsa-live} shows the two failure modes. \textbf{KV-blind prop.} gives the most memory to experts and therefore has the fastest decode, but it leaves only 0.5\,GiB for KV; TTFT rises to 19.8\,s and turn wall reaches 22.0\,s. \textbf{50--50} leaves more KV, but its smaller expert cap slows decode to 29.2 tok/s. \textbf{Naive offload} streams every expert and pays on both axes. \textbf{MV-WSA} balances the two pools: it matches the best TTFT, keeps decode well above 50--50, and has the lowest median turn wall.

\paragraph{The admission floor is what makes the right split feasible.}
The KV-blind prior in Table~\ref{tab:mvwsa-live} shows the soft version of KV starvation: the engine runs, but TTFT collapses. The hard version is a startup failure. If an allocator pushes the split too far toward experts, the remaining KV pool may be too small to admit even one requested context, and the engine can refuse to initialize. The admission floor in Eq.~\eqref{eq:admission} prevents this by clamping the split to the most expert-heavy allocation that still satisfies the scheduler's KV requirement.

\paragraph{Live dual-resize: reclaiming idle KV.}
The closed-form controller of \S\ref{sec:method-dynamic} runs in the in-process engine and resizes both pools at drained barriers. On the 24\,GiB RTX 3090, we replay two Qwen3 AgentInstruct traces (\texttt{os}, \texttt{db}) at concurrency~4. In both traces, the fixed offline split reserves more KV than the workload actually uses: the KV pool peaks at only ${\approx}44\%$ of its allocation each round. The controller observes this after the first drained round and moves the unused KV budget to resident experts (cap $32\!\to\!35$, KV $2.79\!\to\!1.53$\,GiB), then holds that split with zero preemptions.

Table~\ref{tab:mvwsa-dynamic} shows the effect. Dynamic resizing reduces end-to-end time by $1.07$--$1.19\times$ over the fixed offline split at the same GPU budget, and improves TTFT by $1.11$--$1.27\times$. Decode improves on \texttt{db} and is essentially unchanged on \texttt{os}. A separate resize-cycle check confirms property (D2): outputs remain byte-identical across a shrink-KV/grow-experts and grow-KV/shrink-experts cycle. The gain is regime-dependent: when the workload is already KV-pressured, as in Table~\ref{tab:mvwsa-live}, the static configurator is already near the right split and there is little idle KV to reclaim.

\begin{table}[t]
\caption{Live MV-WSA dual-resize (\emph{dynamic}: both pools resized \emph{while} serving) vs.\ the fixed offline split, on a real 24\,GiB RTX 3090, Qwen3-30B-A3B, AgentInstruct~\citep{zeng2024agenttuning} \texttt{os} and \texttt{db}, concurrency~4, 36 timed turns. Both arms replay the byte-identical trace; the only difference is the policy. Lower is better except decode. The dynamic arm reclaims idle KV into experts in one move (cap $32\!\to\!35$, KV $2.79\!\to\!1.53$\,GiB) and reduces end-to-end time on both traces.}

\label{tab:mvwsa-dynamic}
\vskip 0.1in
\begin{center}
\scalebox{0.62}{
\begin{tabular}{@{}llccccc@{}}
\toprule
Trace & Arm & cap & KV pool & TTFT med (s) & decode (tok/s) & e2e (s) \\
\midrule
\texttt{os} & fixed            & 32                 & 2.79\,GiB                   & 89.31 & 3.17 & 1237.1 \\
            & \textbf{dynamic} & 32$\to$\textbf{35} & 2.79$\to$\textbf{1.53}\,GiB & \textbf{80.74} & 3.14 & \textbf{1155.0} \\
            & \emph{dyn/fix}   & ---                & ---                         & $1.11\times$ & $0.99\times$ & $\mathbf{1.07\times}$ \\
\midrule
\texttt{db} & fixed            & 32                 & 2.79\,GiB                   & 95.46 & 3.02 & 1355.4 \\
            & \textbf{dynamic} & 32$\to$\textbf{35} & 2.79$\to$\textbf{1.53}\,GiB & \textbf{75.05} & \textbf{3.36} & \textbf{1138.4} \\
            & \emph{dyn/fix}   & ---                & ---                         & $1.27\times$ & $1.11\times$ & $\mathbf{1.19\times}$ \\
\bottomrule
\end{tabular}
}
\end{center}
\vskip -0.1in
\end{table}

\subsection{Generality across Architectures and Modalities}
\label{sec:eval-general}

The expert-paging mechanism needs only routed experts, so it carries beyond text-only MoE LLMs. On \textbf{Kimi-VL-A3B}, the same vLLM plug-in pages the MoE language backbone with no model-specific code, preserves outputs on both text and image prompts, and improves decode throughput at matched budgets (Section~\ref{sec:eval-isovram}). On the hybrid \textbf{Jamba-v0.1} (52B; Mamba+attention+MoE), WiSP serves the model on one 94\,GiB GPU in a regime where stock vLLM cannot: full residency exceeds VRAM, and the CPU-offload path is unsupported for Jamba's recurrent state. This is a capability claim scoped to the hybrid model.
For MoE diffusion models without an autoregressive KV cache, the full expert--KV allocation problem does not apply, but the pager still does. A roughly 150-line PyTorch wrapper pages \textbf{DiT-MoE} and \textbf{LLaDA-MoE-7B}; generation remains bit-/token-identical to full residency, while resident expert memory drops by up to $4\times$. These models also expose different working-set structures: DiT has a fully activated per-forward stream with little cross-step reuse, while LLaDA has strong temporal locality across denoising steps. Appendix~\ref{app:extended-eval} reports the full mechanism-generalization results (Figures~\ref{fig:ditmoe},~\ref{fig:llada}).

\section{Conclusion}
\label{sec:conclusion}

Serving an MoE on a GPU that cannot keep it resident is a working-set problem: routed expert weights and the KV cache are two memory-demand streams competing for one small store. WiSP realizes this view for the low-concurrency regime. It pages routed experts through a drop-in, byte-identical vLLM plug-in, improves decode throughput by up to $2.0\times$ over static offload at matched VRAM on a real 24\,GiB GPU, and carries the same expert-paging mechanism to a MoE VLM, a 52B hybrid model whose stock vLLM path cannot serve it on one GPU, and two diffusion MoEs as mechanism-generalization checks.

The main lesson is that in single-stream, memory-bound decode, routing prediction is not the speed lever. There is too little compute to hide PCIe transfers, so the routing signal is more useful as a memory signal: it tells us which expert working set is worth keeping resident. That makes allocation the central problem. MV-WSA splits one VRAM budget between resident experts and KV under an admission floor, and the live controller shows that this split can be changed while serving without changing model outputs. Porting the full estimator into the live serve loop, broadening the dynamic controller beyond the current proof of concept, and validating the MiniMax regime on physically smaller GPUs are the next steps.

\bibliography{wisp}

@inproceedings{shazeer2017moe,
  author    = {Shazeer, Noam and Mirhoseini, Azalia and Maziarz, Krzysztof and Davis, Andy and Le, Quoc and Hinton, Geoffrey and Dean, Jeff},
  title     = {Outrageously Large Neural Networks: The Sparsely-Gated {M}ixture-of-{E}xperts Layer},
  booktitle = {International Conference on Learning Representations (ICLR)},
  year      = {2017}
}

@article{fedus2022switch,
  author  = {Fedus, William and Zoph, Barret and Shazeer, Noam},
  title   = {Switch Transformers: Scaling to Trillion Parameter Models with Simple and Efficient Sparsity},
  journal = {Journal of Machine Learning Research (JMLR)},
  volume  = {23},
  pages   = {1--39},
  year    = {2022}
}

@article{jiang2024mixtral,
  author  = {Jiang, Albert Q. and others},
  title   = {Mixtral of Experts},
  journal = {arXiv preprint arXiv:2401.04088},
  year    = {2024}
}

@article{deepseekv3,
  author  = {{DeepSeek-AI}},
  title   = {{DeepSeek-V3} Technical Report},
  journal = {arXiv preprint arXiv:2412.19437},
  year    = {2024}
}

@article{qwen3,
  author  = {{Qwen Team}},
  title   = {{Qwen3} Technical Report},
  journal = {arXiv preprint arXiv:2505.09388},
  year    = {2025}
}

@inproceedings{zeng2024agenttuning,
  title     = {{A}gent{T}uning: Enabling Generalized Agent Abilities for {LLM}s},
  author    = {Zeng, Aohan and Liu, Mingdao and Lu, Rui and Wang, Bowen and Liu, Xiao and Dong, Yuxiao and Tang, Jie},
  booktitle = {Findings of the Association for Computational Linguistics: ACL 2024},
  pages     = {3053--3077},
  year      = {2024},
  address   = {Bangkok, Thailand},
  publisher = {Association for Computational Linguistics},
  doi       = {10.18653/v1/2024.findings-acl.181},
  url       = {https://aclanthology.org/2024.findings-acl.181/}
}

@inproceedings{liu2024agentbench,
  title     = {{AgentBench}: Evaluating {LLM}s as Agents},
  author    = {Liu, Xiao and Yu, Hao and Zhang, Hanchen and Xu, Yifan and Lei, Xuanyu and Lai, Hanyu and Gu, Yu and Ding, Hangliang and Men, Kaiwen and Yang, Kejuan and Zhang, Shudan and Deng, Xiang and Zeng, Aohan and Du, Zhengxiao and Zhang, Chenhui and Shen, Sheng and Zhang, Tianjun and Su, Yu and Sun, Huan and Huang, Minlie and Dong, Yuxiao and Tang, Jie},
  booktitle = {International Conference on Learning Representations (ICLR)},
  year      = {2024},
  url       = {https://openreview.net/forum?id=zAdUB0aCTQ}
}

@inproceedings{kwon2023paged,
  author    = {Kwon, Woosuk and Li, Zhuohan and Zhuang, Siyuan and Sheng, Ying and Zheng, Lianmin and Yu, Cody Hao and Gonzalez, Joseph E. and Zhang, Hao and Stoica, Ion},
  title     = {Efficient Memory Management for Large Language Model Serving with {P}aged{A}ttention},
  booktitle = {ACM Symposium on Operating Systems Principles (SOSP)},
  year      = {2023}
}

@inproceedings{zheng2024sglang,
  author    = {Zheng, Lianmin and others},
  title     = {{SGLang}: Efficient Execution of Structured Language Model Programs},
  booktitle = {Advances in Neural Information Processing Systems (NeurIPS)},
  year      = {2024}
}

@article{denning1968ws,
  author  = {Denning, Peter J.},
  title   = {The Working Set Model for Program Behavior},
  journal = {Communications of the ACM},
  volume  = {11},
  number  = {5},
  pages   = {323--333},
  year    = {1968}
}

@article{denning1970vm,
  author  = {Denning, Peter J.},
  title   = {Virtual Memory},
  journal = {ACM Computing Surveys},
  volume  = {2},
  number  = {3},
  pages   = {153--189},
  year    = {1970}
}

@article{liu2026fluxmoe,
  author  = {Liu, Qingxiu and He, Cyril Y. and Jiang, Hanser and Wang, Zion and Zhao, Alan and Lee, Patrick P. C.},
  title   = {{FluxMoE}: Decoupling Expert Residency for High-Performance {M}o{E} Serving},
  journal = {arXiv preprint arXiv:2604.02715},
  year    = {2026}
}

@inproceedings{cao2025moelightning,
  author    = {Cao, Shiyi and Liu, Shu and Griggs, Tyler and Schafhalter, Peter and Liu, Xiaoxuan and Sheng, Ying and Gonzalez, Joseph E. and Zaharia, Matei and Stoica, Ion},
  title     = {{MoE-Lightning}: High-Throughput {M}o{E} Inference on Memory-Constrained {GPU}s},
  booktitle = {ACM International Conference on Architectural Support for Programming Languages and Operating Systems (ASPLOS)},
  year      = {2025}
}

@inproceedings{du2024sida,
  author    = {Du, Zhixu and others},
  title     = {{SiDA}: Sparsity-Inspired Data-Aware Serving for Efficient and Scalable Large {M}ixture-of-{E}xperts Models},
  booktitle = {Proceedings of Machine Learning and Systems (MLSys)},
  year      = {2024}
}

@article{eliseev2023offload,
  author  = {Eliseev, Artyom and Mazur, Denis},
  title   = {Fast Inference of {M}ixture-of-{E}xperts Language Models with Offloading},
  journal = {arXiv preprint arXiv:2312.17238},
  year    = {2023}
}

@misc{ktransformers,
  author = {{KVCache.AI}},
  title  = {{KTransformers}: A Flexible Framework for Cutting-Edge {LLM} Inference Optimizations},
  howpublished = {\url{https://github.com/kvcache-ai/ktransformers}},
  year   = {2024}
}

@misc{llamacpp,
  author = {Gerganov, Georgi and others},
  title  = {llama.cpp},
  howpublished = {\url{https://github.com/ggml-org/llama.cpp}},
  year   = {2023}
}

@inproceedings{liu2024kivi,
  author    = {Liu, Zirui and Yuan, Jiayi and Jin, Hongye and Zhong, Shaochen and Xu, Zhaozhuo and Braverman, Vladimir and Chen, Beidi and Hu, Xia},
  title     = {{KIVI}: A Tuning-Free Asymmetric 2bit Quantization for {KV} Cache},
  booktitle = {International Conference on Machine Learning (ICML)},
  year      = {2024}
}

@inproceedings{peebles2023dit,
  author    = {Peebles, William and Xie, Saining},
  title     = {Scalable Diffusion Models with Transformers},
  booktitle = {IEEE/CVF International Conference on Computer Vision (ICCV)},
  year      = {2023}
}

@article{fei2024ditmoe,
  author  = {Fei, Zhengcong and Fan, Mingyuan and Yu, Changqian and Li, Debang and Huang, Junshi},
  title   = {Scaling Diffusion Transformers to 16 Billion Parameters ({DiT-MoE})},
  journal = {arXiv preprint arXiv:2407.11633},
  year    = {2024}
}

@article{nie2025llada,
  author  = {Nie, Shen and others},
  title   = {Large Language Diffusion Models ({LLaDA})},
  journal = {arXiv preprint arXiv:2502.09992},
  year    = {2025}
}

@article{zhu2025lladamoe,
  author  = {Zhu, Fengqi and You, Zebin and Xing, Yipeng and Huang, Zenan and others},
  title   = {{LLaDA-MoE}: A Sparse {M}o{E} Diffusion Language Model},
  journal = {arXiv preprint arXiv:2509.24389},
  year    = {2025}
}

@article{wu2024deepseekvl2,
  author  = {Wu, Zhiyu and others},
  title   = {{DeepSeek-VL2}: Mixture-of-Experts Vision-Language Models for Advanced Multimodal Understanding},
  journal = {arXiv preprint arXiv:2412.10302},
  year    = {2024}
}

@article{kimivl2025,
  author  = {{Kimi Team}},
  title   = {{Kimi-VL} Technical Report},
  journal = {arXiv preprint arXiv:2504.07491},
  year    = {2025}
}

@article{lieber2024jamba,
  author  = {Lieber, Opher and others},
  title   = {{Jamba}: A Hybrid Transformer-{M}amba Language Model},
  journal = {arXiv preprint arXiv:2403.19887},
  year    = {2024}
}

@article{anthony2024blackmamba,
  author  = {Anthony, Quentin and Tokpanov, Yury and Glorioso, Paolo and Millidge, Beren},
  title   = {{BlackMamba}: Mixture of Experts for State-Space Models},
  journal = {arXiv preprint arXiv:2402.01771},
  year    = {2024}
}

@article{li2024aria,
  author  = {Li, Dongxu and others},
  title   = {{Aria}: An Open Multimodal Native Mixture-of-Experts Model},
  journal = {arXiv preprint arXiv:2410.05993},
  year    = {2024}
}

@article{moeinfinity,
  author  = {Xue, Leyang and Fu, Yao and Lu, Zhan and Sun, Chuanhao and Mai, Luo and Marina, Mahesh},
  title   = {{MoE-Infinity}: Efficient {MoE} Inference on Personal Machines with Sparsity-Aware Expert Cache},
  journal = {arXiv preprint arXiv:2401.14361},
  year    = {2024}
}

@inproceedings{pregatedmoe,
  author    = {Hwang, Ranggi and Wei, Jianyu and Cao, Shijie and Hwang, Changho and Tang, Xiaohu and Cao, Ting and Yang, Mao},
  title     = {{Pre-gated MoE}: An Algorithm-System Co-Design for Fast and Scalable Mixture-of-Expert Inference},
  booktitle = {ACM/IEEE International Symposium on Computer Architecture (ISCA)},
  year      = {2024}
}

@inproceedings{hybrimoe,
  author    = {Zhong, Shuzhang and Sun, Yanfan and Liang, Ling and Wang, Runsheng and Huang, Ru and Li, Meng},
  title     = {{HybriMoE}: Hybrid {CPU-GPU} Scheduling and Cache Management for Efficient {MoE} Inference},
  booktitle = {ACM/IEEE Design Automation Conference (DAC)},
  year      = {2025}
}

@inproceedings{finemoe,
  author    = {Yu, Hanfei and Cui, Xingqi and Zhang, Hong and Wang, Hao and Wang, Hao},
  title     = {Taming Latency-Memory Trade-Off in {MoE}-Based {LLM} Serving via Fine-Grained Expert Offloading},
  booktitle = {European Conference on Computer Systems (EuroSys)},
  year      = {2026}
}

@article{dymoe,
  author  = {Huang, Yuegui and Fang, Zhiyuan and Luo, Weiqi and Wu, Ruoyu and Chen, Wuhui and Zheng, Zibin},
  title   = {{DyMoE}: Dynamic Expert Orchestration with Mixed-Precision Quantization for Efficient {MoE} Inference on Edge},
  journal = {arXiv preprint arXiv:2603.19172},
  year    = {2026}
}

@article{freetoken,
  author  = {Yang, Shuo and Fan, Xiaoze and Pan, Melissa and Xi, Haocheng and Wang, Zhe and Sun, Shanlin and Keutzer, Kurt and Han, Song and Zaharia, Matei and Xu, Chenfeng and Stoica, Ion},
  title   = {{FreeToken}: Efficient Edge-Native {MoE} Serving with Bandwidth-Adaptive Execution},
  journal = {arXiv preprint arXiv:2608.16157},
  year    = {2026}
}
\bibliographystyle{mlsys2026}

\newpage
\appendix

\section{Related Work}
\label{sec:related}

WiSP's conceptual basis is Denning's working-set model and thrashing
theory~\citep{denning1968ws,denning1970vm}; our contribution is to recognize low-resource MoE serving as a direct instance and use that vocabulary to reason about expert residency, KV capacity, and admission together. 
On the systems side, PagedAttention in vLLM~\citep{kwon2023paged} and RadixAttention in SGLang~\citep{zheng2024sglang}
brought a paging and prefix-sharing discipline to the KV reference stream; we bring the same discipline
to the expert stream and treat the two as co-contending working sets over one store.

A line of batch-oriented systems serves MoEs on memory-constrained GPUs by overlapping transfer with
compute: MoE-Lightning~\citep{cao2025moelightning} pipelines paged weights and reports up to
$10.3\times$ on a T4, SiDA-MoE~\citep{du2024sida} uses an offline hash predictor of active experts (approximate, ${\sim}1\%$ accuracy drop), and
Mixtral-offloading~\citep{eliseev2023offload} streams experts for a single stream. These target high
throughput at large batch, where there is ample compute to hide transfers. A parallel line predicts or
caches experts to cut miss latency rather than stream blindly: MoE-Infinity~\citep{moeinfinity} keeps a
sparsity-aware expert cache driven by activation tracking on personal machines; Pre-gated
MoE~\citep{pregatedmoe} co-designs the model to pre-compute the next layer's experts one step ahead;
HybriMoE~\citep{hybrimoe} splits expert work across CPU and GPU with an impact-driven prefetch;
FineMoE~\citep{finemoe} retrieves per-request expert maps by semantic and trajectory similarity to
guide prefetching; and DyMoE~\citep{dymoe} assigns experts mixed precision (down to skipping) to shrink
edge transfers. Most share WiSP's premise that expert residency is the bottleneck, but spend the routing signal on prefetching or overlap for speed—which Section~\ref{sec:finding} shows is a poor fit for single-stream decode, where no compute hides a transfer. WiSP instead spends the signal on sizing and treats experts and KV as one budget. The closest concurrent work is FluxMoE~\citep{liu2026fluxmoe}, which, like WiSP, builds on vLLM and reclaims expert VRAM for the KV
cache, but adopts a stream-and-evict policy---paging an expert in just before its layer and evicting it
immediately, pipelined with layer compute---to maximize high-batch throughput. WiSP differs in four
ways: it targets the low-concurrency, cache-and-reuse operating point, where Section~\ref{sec:finding}
shows there is no compute to overlap; it ships as a pure plug-in over the unmodified engine with no
virtual-memory machinery; rather than a fixed, overlap-tuned residency plan it \emph{allocates} the
expert$\leftrightarrow$KV split by marginal value (MV-WSA), online and subject to an admission floor,
and physically resizes both pools while serving; and it contributes the working-set framing, the
measurement of when prediction does and does not pay off, and the cross-architecture breadth. The
systems are complementary across the concurrency axis.

Concurrent with this work, FreeToken~\citep{freetoken} co-designs a full edge-native MoE serving
stack---pipelined full-layer expert loading for prefill, semantic-aware state caching for agentic
context edits, and an elastic expert/KV memory mechanism---and, at decode, serves residual expert
misses through a second channel: executing missing experts in place on the CPU, splitting each
layer's misses between PCIe fill and CPU compute in proportion to profiled bandwidths. This adds a
degree of freedom that allocation alone cannot supply under the single-stream PCIe ceiling of
Section~\ref{sec:finding}: where WiSP can only fetch a missing expert, FreeToken may also compute it
where it resides. The two designs are complementary along the allocation axis: FreeToken provides a
runtime mechanism for reconfiguring the expert/KV split but does not specify a policy for choosing
or adapting that split, which is the question MV-WSA answers, and extending MV-WSA's marginal-value
rule to price a second service channel (fetch versus compute) is a natural next step.

Prior work on MoE routing prediction and speculative loading motivates the predictor we build in
Section~\ref{sec:finding}; we evaluate it directly and report a negative speed result for the
low-concurrency regime, which clarifies that the signal is better spent on memory than latency.
KV-cache compression methods such as KIVI~\citep{liu2024kivi} reduce the other side of the joint
working set and compose with WiSP rather than competing with it. Finally, WiSP is evaluated across MoE
LLMs~\citep{shazeer2017moe,fedus2022switch,jiang2024mixtral,deepseekv3,qwen3}, MoE
VLMs~\citep{wu2024deepseekvl2,kimivl2025,li2024aria}, a hybrid
SSM-MoE~\citep{lieber2024jamba,anthony2024blackmamba}, and MoE diffusion models for
images~\citep{peebles2023dit,fei2024ditmoe} and language~\citep{nie2025llada,zhu2025lladamoe}; we use
these only as test subjects and change none of them.

\section{Dynamic Dual-Resize Control Law}
\label{app:dynamic}

Recall $a = L M_e$. At a drained barrier the conserved budget is measured directly,
$B = \kappa b + C a$, and the target sizes $\tilde C, \tilde\kappa$ are
\begin{align}
\tilde\kappa &= \max\bigl(\kappa_{\min},\ \lceil (1+h)\,p \rceil\bigr), \label{eq:kvtgt}\\
\tilde C     &= \mathrm{clip}\bigl(\lfloor (B - \tilde\kappa\, b)/a \rfloor,\ C_{\min},\, C_{\max}\bigr), \label{eq:captgt}\\
\tilde\kappa &\leftarrow \lfloor (B - \tilde C\, a)/b \rfloor, \label{eq:kvreclaim}
\end{align}
where $p$ is the burst's peak KV occupancy (blocks actually used) and $h$ the KV headroom.
Equation~\eqref{eq:kvtgt} promotes the admission floor~\eqref{eq:admission} from a one-shot startup
constraint to a continuously enforced invariant; \eqref{eq:captgt} is the equimarginal ``remainder to
experts'' rule for a KV step curve; and the recompute~\eqref{eq:kvreclaim} reclaims any slack the cap
clamp leaves, so the move never overcommits ($\tilde C\, a + \tilde\kappa\, b \le B$).

The \emph{expert-scratch resize} grows or shrinks the per-layer GPU scratch $S$ to $\tilde C$ slots,
re-paging now-resident experts from $W^{\mathrm{cpu}}$ and dropping victims on a shrink, with
$\pi, \pi^{-1}$ and the LRU ticks resized alongside. The \emph{KV-pool resize} reallocates the worker's
per-layer KV tensors to $\tilde\kappa$ blocks and rebuilds the scheduler's shared block pool in place,
preserving the cached \texttt{null\_block} identity, done only when the pool is fully drained. To keep
the transient footprint at $\max(\text{old},\text{new}) \le B$ rather than their sum, the controller
always frees the shrinking side before growing the other (on an expert grow: shrink KV, then grow
scratch; on a shrink: the reverse). Every move is wrapped so a failed resize is skipped---the engine is
left in its prior, usable state---and a cap dead-zone $\Delta$ suppresses moves smaller than $\Delta$
experts so the split is stable under round-to-round noise.

\section{Implementation Details}
\label{app:impl}

WiSP registers through the standard \texttt{vllm.general\_plugins} entry point, so a single
\texttt{pip install} suffices; the plug-in fires in every vLLM process before any model class is
imported and patches three methods on the fused-MoE module---weight creation, post-load processing, and
the forward. Weight creation allocates the expert masters on pinned host memory (this is what avoids the
load-time GPU spike); post-load processing builds the per-layer $C$-slot scratch and the maps
$\pi, \pi^{-1}$; the forward runs Algorithm~\ref{alg:wisp} and dispatches the unmodified kernel with
$\pi$ as \texttt{expert\_map}. The FP8 block-quantized path pages four tensors per expert (the weights
and their block scales) and forces the Triton backend to bypass DeepGEMM's load-time post-processing,
which assumes full residency. The dynamic controller's two primitives are \texttt{resize\_layer\_caps}
(grow or shrink the expert scratch, re-paging from the host master) and a KV-pool reallocation that
rebuilds the scheduler's block pool in place (Appendix~\ref{app:dynamic}).

\section{Serving Design Space}
\label{app:designspace}

This appendix expands the placement of WiSP among existing systems summarized in
Section~\ref{sec:scope}. Table~\ref{tab:design} compares WiSP against prior autoregressive MoE-LLM
serving systems along the axes that matter for our setting (routing-awareness, byte-identity, delivery,
and target regime), and Table~\ref{tab:stacks} lists the native serving stacks of the non-AR MoE
modalities and how WiSP attaches to each.

\begin{table*}[t]
\caption{AR MoE-LLM serving systems. WiSP is the low-concurrency, drop-in, reuse-caching point in the design space.}
\label{tab:design}
\vskip 0.1in
\begin{center}
\scalebox{0.6}{
\begin{tabular}{llcccl}
\toprule
System & expert mechanism & routing-aware & byte-identical & delivery & regime \\
\midrule
vLLM \texttt{-{}-cpu-offload-gb}~\citep{kwon2023paged} & static, layer-grained & $\times$ & \checkmark & built-in & any (wasteful) \\
ktransformers~\citep{ktransformers} & CPU-side compute (AMX) & $\times$ & \checkmark & engine & Intel CPU+GPU \\
llama.cpp~\citep{llamacpp} & GGUF offload & $\times$ & $\times$ (lossy) & engine & edge \\
MoE-Lightning~\citep{cao2025moelightning} & pipelined paged weights & \checkmark & \checkmark & engine & high-batch \\
SiDA~\citep{du2024sida} & data-aware hash offload & \checkmark & $\times$ (approx) & engine & high-batch \\
FluxMoE~\citep{liu2026fluxmoe} (concurrent) & stream-and-evict + planner & \checkmark & \checkmark & vLLM integ. & high-batch \\
\textbf{WiSP (ours)} & \textbf{cache-and-reuse via \texttt{expert\_map}} & \checkmark & \checkmark & plug-in (entry point) & \textbf{low-concurrency} \\
\bottomrule
\end{tabular}
}
\end{center}
\vskip -0.1in
\end{table*}

\begin{table*}[t]
\caption{Serving stacks for non-AR MoE modalities and how WiSP attaches.}
\label{tab:stacks}
\vskip 0.1in
\begin{center}
\scalebox{0.7}{
\begin{tabular}{lllll}
\toprule
Modality & Representative & Native serving stack & Native expert paging & WiSP integration \\
\midrule
MoE VLM & Kimi-VL, DeepSeek-VL2 & vLLM (multimodal) & none & same engine plug-in \\
Hybrid SSM--MoE & Jamba & vLLM (\texttt{JambaForCausalLM}) & none; offload unsupported & same plug-in \\
MoE image diffusion & DiT-MoE & diffusers / repo (PyTorch) & none & ${\sim}150$-line wrapper \\
MoE diffusion LLM & LLaDA-MoE & HF Transformers / dInfer & none & same wrapper \\
\bottomrule
\end{tabular}
}
\end{center}
\vskip -0.1in
\end{table*}

\section{The Co-Activation Predictor}
\label{app:predictor}

Section~\ref{sec:finding} uses a lightweight routing predictor to test whether expert prediction can hide page faults in low-concurrency decode. The predictor is deliberately simple and online: it needs no gradients, calibration set, or retraining, and it uses only routing decisions that the serving engine already observes.

\paragraph{Construction.}
For each source layer $\ell$ and later target layer $\ell' > \ell$, the predictor keeps a co-activation table
\[
A_{\ell,\ell'}(e,e') ,
\]
which counts how often source expert $e$ at layer $\ell$ and target expert $e'$ at layer $\ell'$ fire in the same forward pass. We normalize by the source marginal count to obtain an empirical conditional probability,
\[
P_{\ell,\ell'}(e' \mid e)
=
\frac{A_{\ell,\ell'}(e,e')}{\sum_j A_{\ell,\ell'}(e,j)} .
\]
At inference time, after the router has selected experts in earlier layers, the predictor scores a candidate expert $e'$ for target layer $\ell'$ by summing the conditional probabilities from the experts already observed:
\[
s_{\ell'}(e') =
\sum_{\ell < \ell'} \gamma^{\ell' - \ell}
\sum_{e \in R_\ell} P_{\ell,\ell'}(e' \mid e),
\]
where $\gamma \in (0,1]$ is a distance-decay factor. The highest-scoring experts are used as the predicted target-layer experts. The table can start empty and be learned online, or it can be initialized from a population trace or a per-user trace.

\paragraph{Why this is a fair prefetch test.}
The predictor is not intended as a new model architecture or a trained drafter. It is a low-overhead test of whether the routing signal already present during serving is useful for prefetching. It is strong enough to make the prefetch hypothesis plausible: on per-user traces it reaches nontrivial next-expert precision, around $0.5$ in our Qwen3 experiments. Thus the negative result in Section~\ref{sec:finding} is not because routing is completely unpredictable. The failure mode is systems-level: even useful predictions compete with demand paging for the same PCIe bandwidth, and at tight caps the extra speculative transfers can evict experts that are still needed.

\paragraph{Using the same signal for memory sizing.}
Although the predictor does not improve single-stream decode speed, it remains useful as a memory signal. A routing trace reveals the support of the expert working set a workload repeatedly revisits. That support can be used to choose an expert cap without sweeping VRAM budgets: a global cap that is safe for all users may waste memory on most of them, while a workload-specific cap can keep the reused experts resident and return the remaining bytes to the KV cache. This is the role of routing information in WiSP: not to hide transfer latency, but to estimate the resident expert set that the allocator should budget for.

\section{Extended Evaluation}
\label{app:extended-eval}

This appendix gives the full versions of the working-set and cross-architecture results that
Section~\ref{sec:eval-general} summarizes.

\subsection{The Working-Set Law, Made Measurable}

\begin{figure*}[t]
\centering
\includegraphics[width=0.92\textwidth]{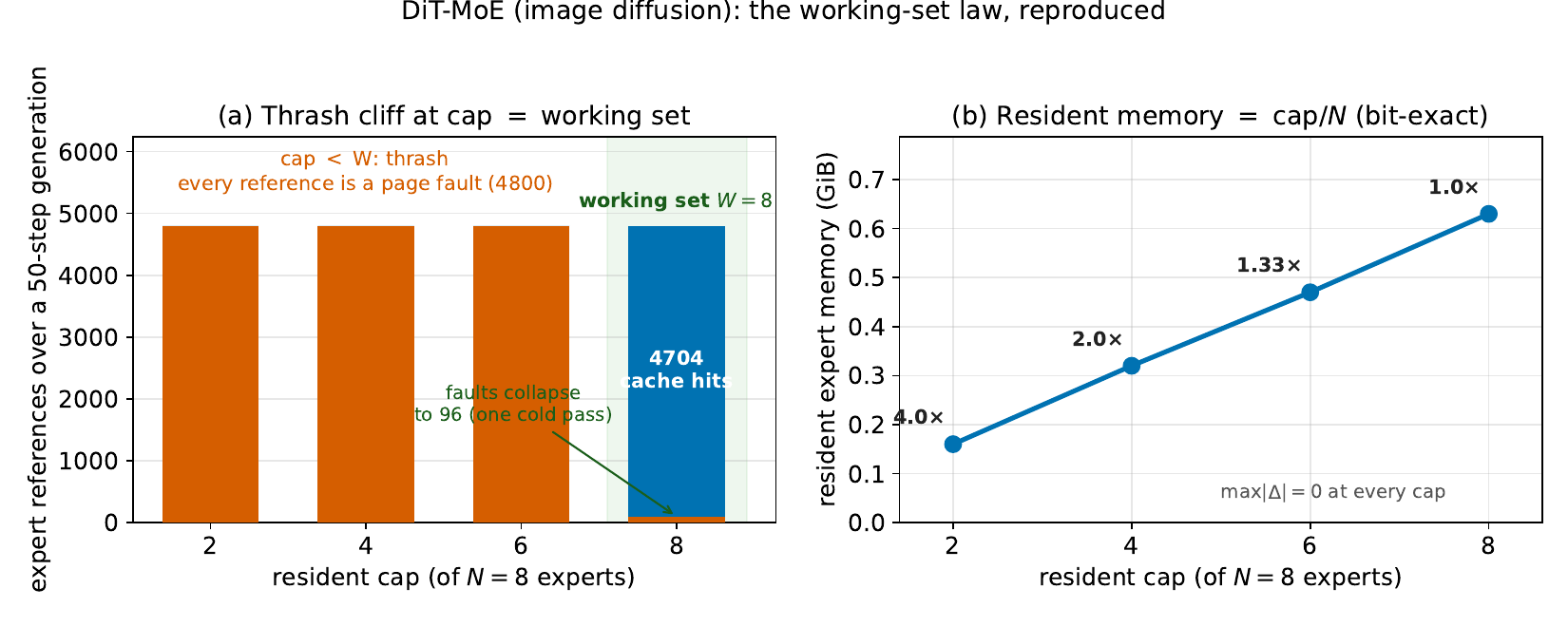}
\caption{Working-set theory reproduced on an image-diffusion MoE (DiT-MoE-S, standalone pager). (a)
Because a diffusion forward activates essentially the whole expert set over its patch tokens, a cap
below the working set thrashes---every reference faults (4800 faults, 0 hits, i.e.\ 8 experts $\times$
12 MoE blocks $\times$ 50 denoising steps)---and at cap = working set the faults collapse to a single
cold pass (96) with 4704 hits. (b) Resident expert memory scales as cap/$N$, up to $4\times$ smaller,
and the generation is bit-exact at every cap.}
\label{fig:ditmoe}
\end{figure*}

Figure~\ref{fig:ditmoe} characterizes the structure that governs where prediction \emph{could} pay off,
on a model where the working set is unusually clean to read off. A diffusion forward routes hundreds of
patch tokens through the experts, so its per-forward working set is essentially the entire expert set,
and the fault counts trace Denning's law exactly: below the working set the cache thrashes, and at the
working set faults collapse and reuse takes over. Resident memory scales linearly with the cap, four
times smaller at the tightest setting, and the output is bit-exact throughout. This is the cleanest
illustration we have of \emph{why} a single, fully-activated reference stream lets paging bound memory
but not latency---and, read the other way, of why diffusion's abundant per-step compute is a regime where a prefetcher would at least have computation to hide behind, unlike single-stream LLM decode. On a fine-grained, heavily load-balanced LLM under diverse prompts (Qwen3-30B-A3B), by contrast, routing is too uniform for right-sizing to exploit and the throughput-versus-cap curve is
essentially topic-independent---a conservative lower bound we report for honesty.

\subsection{A Different Structure: Temporal Locality in a Diffusion LLM}

\begin{figure*}[t]
\centering
\includegraphics[width=0.92\textwidth]{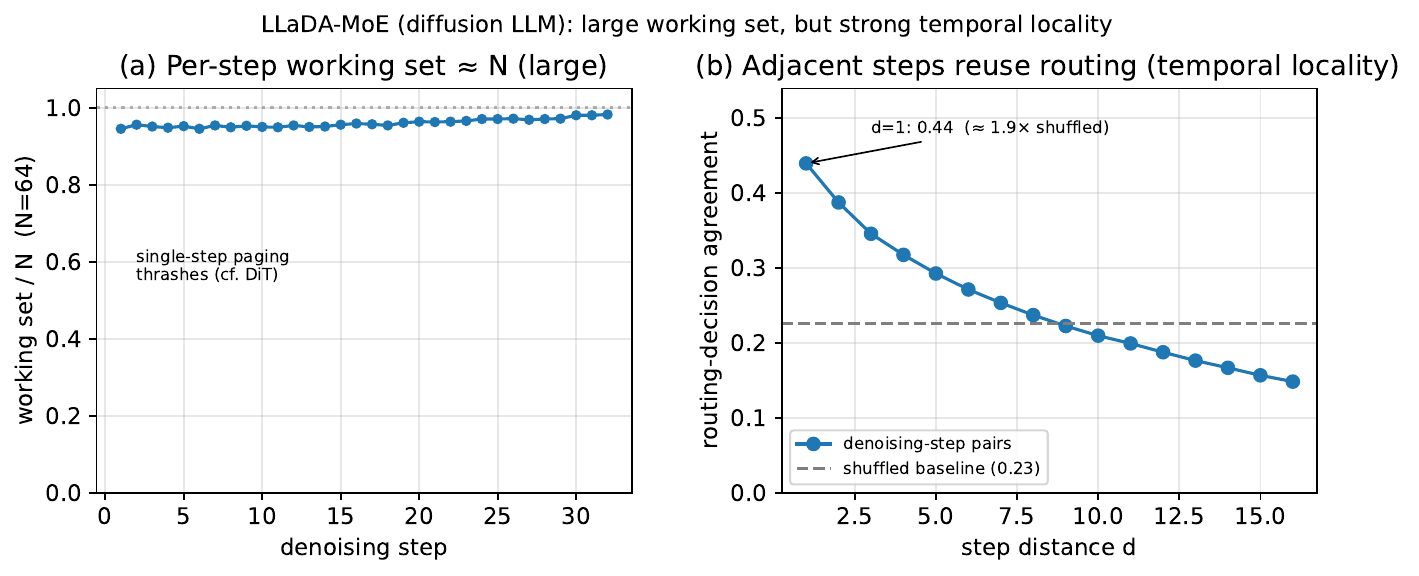}
\caption{LLaDA-MoE has a large but temporally local working set. (a) Each denoising step activates
${\approx}96\%$ of all 64 experts, so---exactly as in DiT---paging a single step thrashes unless the
cap is near $N$. (b) Yet adjacent denoising steps reuse their routing decisions: the per-position
top-$k$ agreement is 0.44 at distance one (${\approx}1.9\times$ a shuffled-step baseline of 0.23) and
decays smoothly with step distance. The structure is temporal, not spatial.}
\label{fig:llada}
\end{figure*}

DiT-MoE shows one kind of working-set structure---spatial, within a single forward, with no reuse below
the working set. A diffusion \emph{language} model exhibits the complementary kind. LLaDA-MoE generates
by iteratively denoising the \emph{same} sequence over many steps, so we ask not how a single forward
routes but how routing evolves \emph{across} denoising steps. Figure~\ref{fig:llada} makes the contrast
with DiT precise. The per-step working set is large---about 96\% of the 64 experts fire in every
step---so a naive single-step pager thrashes just as it does on DiT, and bounding memory is again the
only thing a small cap buys \emph{within} a step. But across steps the picture changes: adjacent
denoising steps agree on 44\% of their per-position routing decisions, nearly twice the 23\% of a
shuffled-order baseline, and the agreement decays monotonically with step distance---the signature of
genuine temporal locality rather than uniform reuse. DiT's structure is spatial and confined to one
forward, so a prefetcher has nothing to anticipate across forwards; LLaDA's structure is temporal, so
the experts a step will need are largely the ones the previous step just used, and a warm or predictive
cache that survives across steps has real structure to exploit. Together these bracket the two regimes
the framing predicts: a fully-activated stream with no cross-step reuse (DiT), where paging is a pure
memory lever, and a fully-activated stream \emph{with} cross-step reuse (LLaDA), where the routing signal has structure a cache or prefetcher could exploit—the diffusion counterpart of the multi-turn agent reuse we leave to future work.

\section{Limitations}
\label{sec:limitations}

WiSP currently runs in eager mode because the plug-in does not yet support CUDA graphs; both sides of our comparisons use eager mode for fairness. The per-layer paging decision also sits on the Python hot path. This overhead matters less in the sub-fitting regime where paging avoids large PCIe transfers, but it becomes visible as the expert cap grows and there is less to page. Moving this decision off the Python path is the most direct systems optimization. We also do not benchmark ktransformers on its Intel AMX fast path, so our comparison to that system is a regime disclosure rather than a measured loss.

Our headline iso-VRAM results for Qwen3 and Kimi-VL are measured on a physically constrained 24\,GiB RTX 3090 (PCIe 4.0). This removes the emulation issue, but it also means the card cannot reach the full-footprint budget where static residency becomes faster; we characterize that crossover only on a larger GPU. The MiniMax-M2 result is likewise scoped as a capability check, not an iso-throughput win.

The MV-WSA results are also scoped. The startup-configurator result in Table~\ref{tab:mvwsa-live} is a real \texttt{vllm serve} run, but it covers one KV-pressured setting. The live dual-resize controller in Table~\ref{tab:mvwsa-dynamic} is a proof of concept on real kernels: it uses the closed-form step-curve rule rather than the full equimarginal estimator, runs inside the in-process v1 engine rather than the multi-process \texttt{vllm serve} loop, and is evaluated on one model family and two workloads. Its byte-identity under resizing is checked by a dedicated resize-cycle test rather than a continuous regression suite. Broader workload coverage, serve-loop integration, self-calibrated cost constants, and porting the full estimator onto the live resize path are the next engineering steps.

\end{document}